\documentclass[letterpaper]{article}
\usepackage[preprint]{aaai2027}
\usepackage[hyphens]{url}
\usepackage{graphicx}
\usepackage{natbib}
\usepackage{caption}
\usepackage{amsmath}
\usepackage{amssymb}
\usepackage{booktabs}
\usepackage{colortbl}

\definecolor{softblue}{RGB}{232,244,255}
\definecolor{softred}{RGB}{255,235,235}
\definecolor{softorange}{RGB}{255,242,224}
\definecolor{tabbest}{RGB}{176,32,32}
\definecolor{tabsecond}{RGB}{32,80,176}
\newcommand{\methodname}{SPARED}

\title{SPARED: Reasoning-Based AI-Generated Image Detection via Adversarially Edited Data}
\author{
Yicheng Bao\textsuperscript{\rm 1},
Xiahui Guo\textsuperscript{\rm 1},
Xuhong Wang\textsuperscript{\rm 2,*},
Xin Tan\textsuperscript{\rm 1,2,*}
}
\affiliations{
\textsuperscript{\rm 1}East China Normal University\\
\textsuperscript{\rm 2}Shanghai Artificial Intelligence Laboratory\\
\textsuperscript{*}Co-corresponding authors
}

\begin{document}

\maketitle

\begin{abstract}
Detecting AI-generated images is only half the task: a deployed detector must also justify its verdict, yet existing detectors inherit three failure modes from their training data: real and fake images collected from different sources invite provenance shortcuts, supervised explanation corpora teach templated rationales, and a static forgery corpus leaves the decision boundary standing still while generators keep moving. We introduce \methodname{}, an adversarial reinforcement learning framework that pits two heterogeneous models against each other. A diffusion image editor learns to edit real photographs into fake counterparts of those same photographs that fool the current detector, while a reasoning MLLM learns to expose them with a verdict grounded in free-form reasoning. Both rewards are shortcut-proof by design: the attacker is credited only when its edit is faithfully executed, and the defender only when its verdict is correct. As the two models alternate, each round's attacker regenerates a harder training pool aimed at the current detector's blind spots, so the detector must generalize rather than memorize any fixed artifact distribution. Although the explanation is never rewarded, its quality rises round over round as a side effect of accuracy-only training. A detector trained within this loop improves monotonically across rounds on each of three external benchmarks.
\end{abstract}

\section{Introduction}

\begin{figure}[t!]
\centering
\includegraphics[width=\columnwidth]{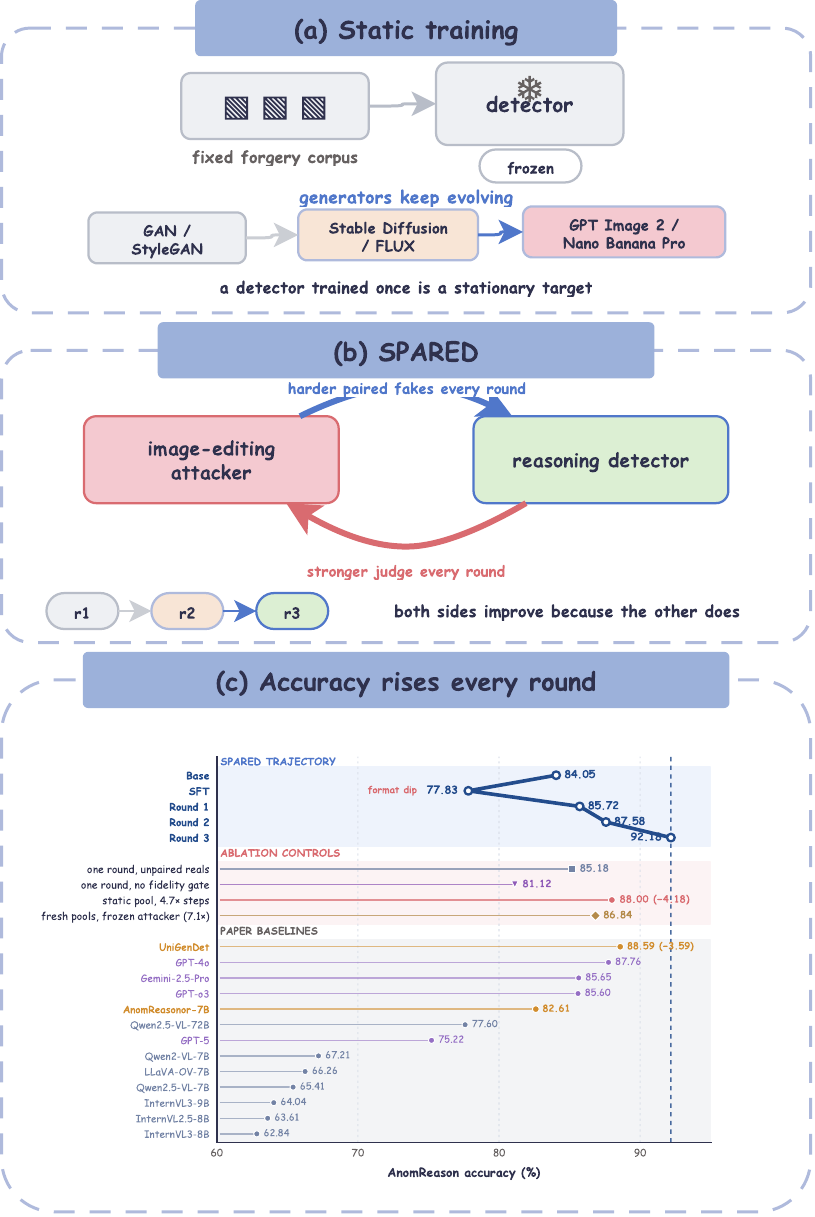}
\caption{(a) A detector trained once on a static corpus is a stationary target for generators that keep moving. (b) \methodname{} instead trains an image-editing attacker against a reasoning detector in alternation: each round, the attacker forges harder paired fakes and the retrained detector becomes a stronger judge. (c) The same 9B backbone, trained in this loop, improves round over round on three external benchmarks; the SFT dip trades accuracy for the reasoning output format, which the adversarial rounds more than recover.}
\label{fig:teaser}
\end{figure}

Generalizable, explainable AI-generated image (AIGI) detection takes a single image (a real camera photograph, a fully AI-generated image, or a locally AI-edited photograph) and outputs a real/fake verdict together with a natural-language account of which regions or cues make the image look synthetic. This judge-style formulation, exemplified by recent MLLM-as-a-judge deepfake benchmarks~\citep{Kuckreja2026Pixels} and explainable forensics models~\citep{Zhou2025Aigiholmes,Huang2025Sida}, matters for three reasons. First, moderation must both flag a fabricated image and justify the flag: an unexplained label is easy to dismiss, and a wrong flag on a realistic photo can itself trigger misinformation. Second, deployed detectors meet fakes from an open, growing population of generators~\citep{Park2025Community,Huang2026Mirage}, so what decides practical value is generalization beyond a fixed forgery distribution, not accuracy on it. Third, public detectors can be targeted by successful white- and black-box evasion attacks~\citep{Diao2024Vulnerabilities}, so a detector that cannot keep improving after deployment is already obsolete.

This problem is hard because three generations of detectors have already failed it, for different reasons. Traditional signal-, frequency-, and pixel-level detectors~\citep{Frank2020Leveraging,Wang2020Cnngenerated,Tan2023Rethinking} identify upsampling fingerprints, spectral artifacts, or neighboring-pixel irregularities left by a specific family of generator architectures. They can perform well in-distribution but degrade sharply when the fingerprint changes: spectral artifacts differ substantially across generative models, hindering generalization to unseen generators~\citep{Karageorgiou2025Anyresolution}. Recent generators such as FLUX~\citep{blackforestlabs2024flux1dev} and Stable Diffusion 3.5~\citep{esser2024scaling,stabilityai2024stable35large} further broaden the synthesis pipelines these detectors must cover. Robust detection therefore also needs semantic evidence: physically implausible geometry, broken object counts, and commonsense violations that require world knowledge rather than pixel-level pattern matching to notice~\citep{Tan2025Semantic,Kim2026Eyes}.

To capture this semantic residue, a second line of work fine-tunes large pretrained models (CLIP-based detectors~\citep{radford2021learning,Ojha2023Towards} or MLLMs) on labeled real/fake data, but it fails along an orthogonal axis. Debiasing efforts~\citep{Guillaro2025Biasfree,Rajan2025Aligned,Chen2025Dual} and the sanity-check literature~\citep{Yan2024Sanity} have identified a dominant source of inflated generalization numbers. When real and fake images in a training set differ in resolution, JPEG quality, or semantic content (because they were sourced independently rather than paired), the detector learns to key on this dataset-level style gap instead of the generation trace itself. Its accuracy then collapses on any semantically aligned, unbiased test set. The result is a field where reported generalization gains are hard to disentangle from residual dataset bias.

A third, more recent line reframes detection as MLLM reasoning to inherit world knowledge and produce genuine explanations~\citep{Zhou2025Aigiholmes,Tan2025Semantic,Huang2025ThinkFake,Li2025Raidx,Ji2026FakeXplain}, but two further failure modes emerge here. Supervised fine-tuning on a fixed explanation corpus tends to produce templated, causally shallow rationales rather than an understanding that transfers to unseen artifacts~\citep{Zhou2025Aigiholmes}; and because the training corpus, however well debiased, is still static, the resulting detector is a stationary target: a fixed decision boundary that the next generator release can be tuned against. The fakes inside a static dataset are also, by construction, only as hard as the detector that existed when the dataset was built. Single-model self-evolution attempts such as ForeAgent~\citep{Wu2026Perception} revise their own reasoning traces on this same fixed data and inherit the same ceiling: no external adversary manufactures new hard cases. The same structural failure, a static corpus permanently behind an adaptive adversary, has already been addressed with attacker-defender training loops in LLM safety~\citep{Dai2025Secure,Wen2026Magic,Liu2025Chasing} and reasoning~\citep{Huang2025Rzero}, and in face-forgery detection with synthesizers that adapt their forgeries to the current detector~\citep{Chen2022Selfsupervised,Lin2024Fake,Zhou2026Crda}. That literature also records the loop's own characteristic failure mode: an adversary rewarded only for winning drifts toward degenerate wins (trivial or off-distribution attacks) unless its reward forces genuine engagement.

We bring this paradigm into visual forgery detection as \methodname{} (\textbf{S}hortcut-\textbf{P}roof \textbf{A}dversarial \textbf{R}easoning over \textbf{E}dited \textbf{D}ata; Figure~\ref{fig:teaser}), with a generative image editor as the adversary rather than a policy over a fixed pool of forgery operations. We find that it also closes the dataset-bias shortcut diagnosed above, as a side effect of how its training data is constructed. Our defender is a reasoning MLLM built on a Qwen3.5-9B backbone~\citep{qwen2026qwen35} and trained with accuracy-based GRPO~\citep{shao2024deepseekmath} to output a free-form reasoning trace followed by an explicit verdict. Because the verdict is the only signal rewarded, explanations cannot score by reproducing templates and improve only insofar as they lead to correct verdicts. This answers the templated-rationale failure above. Our attacker is an RL-optimized image-editing model, a Qwen-Image-Edit-2511 LoRA~\citep{wu2025qwenimagetechnicalreport} trained with DiffusionNFT~\citep{zheng2025diffusionnft}, and it does not synthesize fakes from scratch: it edits real source photographs into their fake counterparts. Because every fake is produced from its own real source image, the two share near-identical content, composition, and resolution, and the only systematic difference left for the defender to learn from is the residual editing trace itself. This closes the content/format shortcut identified above~\citep{Guillaro2025Biasfree} by construction, rather than by post-hoc alignment.

\begin{figure*}[t!]
\centering
\includegraphics[width=\textwidth]{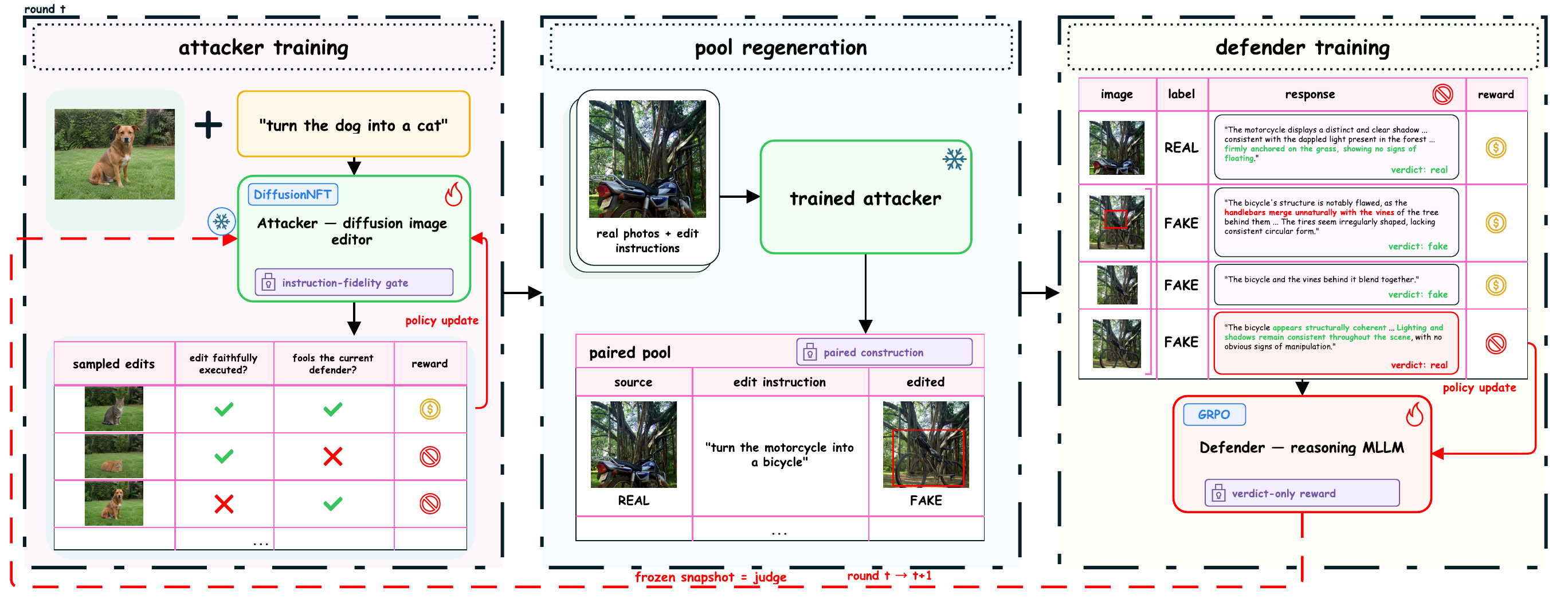}
\caption{Overview of \methodname{}: one adversarial round, read left to right. \textbf{Attacker training}: from one source photograph and its edit instruction the attacker samples a group of candidate edits, and the gated reward credits a rollout only when the instruction was faithfully executed \emph{and} the result fools the frozen defender, so fooling without editing earns nothing (lock~2). \textbf{Pool regeneration}: the trained attacker re-edits a fixed set of real photographs into pairs whose two halves differ only by the edit (lock~3); the pair shown is an actual training pair. \textbf{Defender training}: both halves of a pair are supervised, and among the sampled responses only verdict correctness is rewarded, so explanations cannot be gamed (lock~1). The retrained defender's frozen snapshot becomes the judge for round $t{+}1$.}
\label{fig:method}
\end{figure*}

Closing the dataset-bias shortcut this way only works if the attacker keeps editing in earnest, so the second half of our design is the reward that keeps it honest. The attacker is trained with a gated reward: it is credited only when an instruction-following gate (PaCo)~\citep{ping2025pacorl} confirms the edit was faithfully executed and the resulting pair defeats the current defender. The game therefore cannot be won by skipping the edit or degenerating into off-manifold noise. This gate is the vision-domain analogue of the anti-collapse mechanisms other self-play systems require~\citep{Dai2025Secure,Wen2026Magic,Huang2025Rzero}. The result is an automatically escalating curriculum of instruction-faithful hard negatives, aimed at the current defender's blind spot, that no fixed dataset can provide.

The two models are also architecturally heterogeneous, a diffusion editor in continuous pixel space against an autoregressive MLLM over discrete tokens, and share no parameters. This is a stronger form of the decoupling that MAGIC~\citep{Wen2026Magic} shows avoids the gradient conflict of shared-parameter self-play, and it contrasts with the closest unified systems in this domain, which couple generation and detection~\citep{Zhang2026Unigendet} or correction and detection~\citep{Xu2026Genshield} inside one backbone.
Trained on this self-generated data beyond a lightweight SFT initialization, \methodname{} improves monotonically over three adversarial iterations across three external benchmarks, two of them fully zero-shot (Tables~\ref{tab:detect_revised} and~\ref{tab:anomreason_deepfake_results}, Figure~\ref{fig:analysis}). On judge-style detection~\citep{Kuckreja2026Pixels} and semantic-anomaly reasoning~\citep{Tan2025Semantic} it surpasses every baseline we evaluate, with the single exception of a reasoning model 26 times its size on the former, and its explanation quality rises together with accuracy although the explanation is never rewarded. It also transfers to the ten unseen generator families of Holmes-Set~\citep{Zhou2025Aigiholmes}, whose fully synthetic images are unlike the locally edited photographs it trains on.

Our contributions are:
\begin{itemize}
\item A decoupled attacker--defender reinforcement-learning loop for visual forgery detection whose adversary is a generative image editor rather than a policy over a fixed pool of forgery operations, showing that editing real images into semantically aligned fakes closes this field's dataset-bias shortcut at the training-data level, rather than through post-hoc alignment.
\item A heterogeneous, continuous-pixel-space adversary (a PaCo-gated, RL-trained diffusion image editor) paired with a decoupled reasoning-MLLM defender, extending the anti-collapse mechanisms of prior self-play systems to the vision domain.
\item Empirical evidence that adversarially regenerated edited data alone induces monotonic generalization across three external benchmarks (two of them zero-shot) spanning judge-style detection, semantic-anomaly reasoning, and ten unseen image generators, together with an explicit account of where this signal regresses.
\end{itemize}

\section{Method}

\subsection{Overview}

Our pipeline instantiates the attacker-defender loop introduced in the Introduction (Figure~\ref{fig:method}) as two independently parameterized models trained in alternation: a reasoning defender ($\pi_D$) that maps an image to a verdict and an explanation, and an image-editing attacker ($\pi_A$) that maps a real photograph and an edit instruction to a synthetic counterpart of that same photograph. Round $t$ trains $\pi_A$ against a frozen snapshot of $\pi_D^{(t)}$ to produce a new pool of hard-negative edited images, and the next round trains $\pi_D$ on that same pool; the two models never share parameters or a training objective, only this exchanged data and reward signal.

Every design decision below follows a single principle: \emph{in an adversarial training loop, any reward or data channel that can be satisfied by a shortcut will be, so the loop escalates genuine capability only if every channel connecting the two models is shortcut-proof.} The loop has exactly three such channels, each admitting one of the failure modes diagnosed in the Introduction, and each subsection below closes one: the defender's reward credits only the final verdict, so explanations cannot be gamed independently of detection (Reasoning Defender); the attacker's reward is gated on instruction fidelity, so fooling without genuinely editing earns nothing (PaCo-Gated Diffusion Attacker); and every fake is paired with its own real source under strict deduplication, so provenance cannot substitute for the editing trace (Data Construction and Training Schedule). With all three channels closed, escalation can occur only through capability: the data gets harder only if the attacker genuinely improves, and the defender gains only by defeating data constructed to defeat it.

\subsection{Reasoning Defender}

On the defender's channel, the shortcut is the explanation itself: a detector supervised to reproduce fixed explanation text can satisfy its objective by imitating the phrasing of its training rationales, without ever locating the causal evidence behind a verdict. We therefore leave the defender's reasoning trace free-form and optimize only the final verdict, so that any reasoning strategy the model discovers is retained solely because it correlates with getting the label right.

The defender is built on a Qwen3.5-9B multimodal backbone~\citep{qwen2026qwen35}. Given an image $x$, it generates a response containing an open-ended \texttt{<reasoning>}...\texttt{</reasoning>} span followed by \texttt{<answer>}$\hat{y}$\texttt{</answer>}, where $\hat{y}\in\{\text{real}, \text{fake}\}$. We first LoRA-tune~\citep{hu2022lora} the base model on real/fake reasoning pairs from the judge-training corpus released with DeepfakeJudge~\citep{Kuckreja2026Pixels}, disjoint from its evaluation split, to teach the tag format and elementary artifact vocabulary, then merge the adapter and continue with full-parameter GRPO~\citep{shao2024deepseekmath}. The corpus size and the adapter and optimization hyperparameters are listed in the technical appendix. For each training image $x$ with ground-truth label $y$, we sample a group of $G$ responses and assign
\begin{equation}
r_D(x,y) = \mathbb{1}[\hat{y} = y],
\end{equation}
crediting nothing for an unparsable answer or a wrong label; GRPO normalizes $r_D$ within the group to form the policy-gradient advantage, with no separate reward term for format, length, or reasoning content.

Because $r_D$ never rewards the reasoning span directly, reasoning is reinforced only through the correct verdicts it leads to: explanation quality becomes an emergent side effect of accuracy rather than an independent target that can be gamed on its own.

\subsection{PaCo-Gated Diffusion Attacker}

On the attacker's channel, the shortcut is fooling without editing: were $\pi_A$ rewarded purely for defeating $\pi_D$, the optimum would be degenerate, since leaving the image essentially unedited and pushing it off the manifold of plausible photographs both fool a detector without producing a meaningful hard negative. Closing this shortcut means making a faithfully executed edit a precondition for any fooling credit.

$\pi_A$ is a LoRA adapter~\citep{hu2022lora} on Qwen-Image-Edit-2511~\citep{wu2025qwenimagetechnicalreport}, trained online with DiffusionNFT~\citep{zheng2025diffusionnft}. Given a real source photograph $x_{\text{src}}$ and an edit instruction $c$, both drawn from a mixed single-turn editing corpus (ImgEdit, pico-banana-400k, MagicBrush)~\citep{ye2025imgedit,qian2025picobanana,zhang2023magicbrush}, $\pi_A$ produces an edited image $x_{\text{edit}} = \pi_A(x_{\text{src}}, c)$, paired with its own source as $(x_{\text{src}}, \text{real})$ and $(x_{\text{edit}}, \text{fake})$. Every rollout is scored by two frozen judges: an instruction-following scorer $r_{\text{PaCo}}(x_{\text{src}}, c, x_{\text{edit}}) \in [0,1]$~\citep{ping2025pacorl}, and the current defender snapshot $\pi_D^{(t)}$, queried on both images in the pair to produce a reversed detection reward
\begin{equation}
r_{\text{det}} = \mathbb{1}\big[\lnot(\pi_D^{(t)}(x_{\text{src}}){=}\text{real} \,\wedge\, \pi_D^{(t)}(x_{\text{edit}}){=}\text{fake})\big],
\end{equation}
i.e., the attacker earns nothing when the defender gets the whole pair right. The reward is pair-level: the attacker is also credited when the defender misjudges the untouched source, which steers the regenerated pool toward the defender's real-side errors as well as its fake-side ones. The final reward gates the second on the first:
\begin{equation}
r_A = \begin{cases} r_{\text{det}} & \text{if } r_{\text{PaCo}} \ge 0.7, \\ 0 & \text{otherwise,} \end{cases}
\end{equation}
optimized with DiffusionNFT.

The gate makes $r_{\text{det}}=1$ a necessary but not sufficient condition for reward: an edit that fools $\pi_D^{(t)}$ by ignoring the instruction still scores zero, so only rollouts that are simultaneously instruction-faithful and adversarial to the defender's current boundary receive positive reward. Because $\pi_D^{(t)}$ is refreshed every round, the region of image space that satisfies both constraints moves with the defender, so each round's data pool arrives at a difficulty level calibrated to the model it will retrain.

\subsection{Data Construction and Training Schedule}

The third channel is the training pool itself. Its shortcut, separating reals from fakes by provenance rather than by editing trace, is a property of how the pool is assembled rather than of either model, so it must be closed at the data-construction stage. Every fake in every round must come from editing a real image that is also present as that pair's real label, and no source image may leak into the images used to build our evaluation benchmarks.

Both the real sources and the edit instructions come from the mixed single-turn editing corpus introduced above. A fixed source set is randomly sampled from this corpus once, deduplicated, and screened by perceptual-hash matching against every benchmark used in our subsequent evaluation, so that no training source leaks into any reported result. Each round's pool then re-edits this same source set with the current attacker; corpus filtering rules and screening thresholds are given in the technical appendix.

The resulting schedule alternates the two models over five rounds. Starting from the LoRA-SFT checkpoint merged into a full model, the first defender GRPO round (Iter1) trains on a pool edited by the base image-editing model. Each attacker round (A1, A2) then trains $\pi_A$ against a frozen snapshot of the preceding defender, and the next defender round (Iter2, Iter3) continues GRPO from the previous defender checkpoint on a pool regenerated by the newly trained attacker; Iter3 is the defender reported as \methodname{} throughout this paper.

Alternating rather than jointly optimizing the two models keeps each optimization well-posed: $\pi_A$ needs a judge that is stationary within an RL run, and $\pi_D$ needs a pool that does not shift mid-run.

\section{Experiments}

\begin{table}[t!]
\centering
\renewcommand{\arraystretch}{1.0}
\resizebox{\columnwidth}{!}{
\begin{tabular}{lcccccc}
\toprule
\textbf{Model} & \textbf{Real Acc} & \textbf{Real F1} & \textbf{Fake Acc} & \textbf{Fake F1} & \textbf{Overall Acc} & \textbf{Overall F1} \\
\midrule
\multicolumn{7}{c}{\textit{Open}} \\
\midrule
InternVL3.5-1B-HF~\citep{wang2025internvl35} & 47.8 & 63.0 & 47.8 & 11.4 & 47.8 & 37.2 \\
Qwen3-VL-2B-Instruct~\citep{bai2025qwen3vl} & 49.8 & 44.7 & 49.8 & 54.0 & 49.8 & 49.4 \\
Qwen3-VL-8B-Instruct~\citep{bai2025qwen3vl} & 50.4 & 23.7 & 50.4 & 63.2 & 50.4 & 43.5 \\
Google-Gemma-12B~\citep{gemmateam2025gemma3} & 57.7 & 57.4 & 49.4 & 54.0 & 53.6 & 55.7 \\
Microsoft-Phi-4-Instruct~\citep{abdin2024phi4} & 61.0 & 60.8 & 54.4 & 58.2 & 57.7 & 59.5 \\
InternVL3.5-GPT-OSS-20B-A4B~\citep{wang2025internvl35} & 55.6 & 67.6 & 55.6 & 29.2 & 55.6 & 48.4 \\
Qwen3-VL-30B~\citep{bai2025qwen3vl} & 94.6 & 74.5 & 41.0 & 56.0 & 67.8 & 65.3 \\
Qwen3-VL-235B~\citep{bai2025qwen3vl} & 93.5 & 78.6 & 55.4 & 68.4 & 74.5 & 73.5 \\
\midrule
\multicolumn{7}{c}{\textit{Closed}} \\
\midrule
Gemini-2.5-Flash~\citep{comanici2025gemini25} & 96.6 & 73.7 & 34.5 & 50.0 & 65.6 & 61.9 \\
ChatGPT-4o-mini~\citep{openai2024gpt4omini} & 95.8 & 70.2 & 22.7 & 35.8 & 59.3 & 53.0 \\
\midrule
\multicolumn{7}{c}{\textit{Reasoning}} \\
\midrule
Qwen3-VL-8B-Thinking~\citep{bai2025qwen3vl} & 67.1 & 67.1 & 78.7 & 69.9 & 72.9 & 68.5 \\
Qwen3-VL-30B-Thinking~\citep{bai2025qwen3vl} & 67.4 & 66.0 & 87.6 & 72.9 & 77.5 & 69.5 \\
Qwen3-VL-235B-Thinking~\citep{bai2025qwen3vl} & 75.0 & 76.6 & 90.3 & 79.8 & \textcolor{tabbest}{\textbf{82.7}} & \textcolor{tabsecond}{\underline{78.2}} \\
\midrule
\multicolumn{7}{c}{\textit{Deepfake-Specialized}} \\
\midrule
SIDA-13B-Description~\citep{Huang2025Sida} & 67.6 & 57.0 & 27.9 & 34.5 & 47.8 & 45.8 \\
Qwen2.5-VL-Gen-Buster++~\citep{wen2025busterxpp} & 49.9 & 40.0 & 49.9 & 66.5 & 49.9 & 53.3 \\
UniGenDet~\citep{Zhang2026Unigendet} & 0.0 & 0.0 & 99.9 & 66.6 & 50.0 & 33.3 \\
\midrule
\multicolumn{7}{c}{\textit{Ours}} \\
\midrule
Qwen3.5-9B~\citep{qwen2026qwen35} & 65.7 & 68.7 & 65.0 & 72.1 & 65.4 & 70.4 \\
Qwen3.5-9B-LoRA-SFT~\citep{qwen2026qwen35,hu2022lora} & 57.1 & 65.4 & 82.0 & 73.0 & 69.6 & 69.2 \\
Qwen3.5-9B-Iter1~\citep{qwen2026qwen35} & 59.3 & 68.2 & 84.9 & 75.3 & 72.1 & 71.8 \\
Qwen3.5-9B-Iter2~\citep{qwen2026qwen35} & 66.4 & 73.5 & 85.5 & 78.1 & 76.0 & 75.8 \\
\rowcolor{softblue}
Qwen3.5-9B-Iter3~\citep{qwen2026qwen35} & 70.2 & 77.4 & 88.8 & 81.2 & \textcolor{tabsecond}{\underline{79.5}} & \textcolor{tabbest}{\textbf{79.3}} \\
\bottomrule
\end{tabular}
}
\caption{Comparison of SOTA open-source, closed-source, reasoning, and deepfake models on the DeepfakeJudge-Detect dataset~\citep{Kuckreja2026Pixels}. Invalid or unparsable answers are counted as incorrect; Real Acc and Fake Acc are per-class recall, and Overall F1 is the macro average of real and fake F1. Red bold and blue underline mark the best and second-best overall scores; the shaded row is our final model.}
\label{tab:detect_revised}
\end{table}

\begin{table}[t!]
\centering
\renewcommand{\arraystretch}{1.0}
\resizebox{\columnwidth}{!}{
\begin{tabular}{lcccc}
\toprule
\textbf{Model} & \textbf{Acc} & \textbf{CSemAP-Phe} & \textbf{CSemAP-Rea} & \textbf{CSemAP-Full} \\
\midrule
\multicolumn{5}{c}{\textit{Open}} \\
\midrule
LLaVA-OV-7B~\citep{li2024llavaonevision} & 66.26 & 0.1235 & 0.1124 & 0.1141 \\
Phi-3.5-Vision~\citep{abdin2024phi3} & 41.33 & 0.0685 & 0.0602 & 0.0616 \\
InternVL2.5-8B~\citep{chen2024internvl25} & 63.61 & 0.1165 & 0.1085 & 0.1089 \\
InternVL3-8B~\citep{zhu2025internvl3} & 62.84 & 0.2949 & 0.2394 & 0.2559 \\
InternVL3-9B~\citep{zhu2025internvl3} & 64.04 & 0.2293 & 0.1987 & 0.2081 \\
Qwen2-VL-7B~\citep{wang2024qwen2vl} & 67.21 & 0.1710 & 0.1421 & 0.1496 \\
Qwen2.5-VL-7B~\citep{bai2025qwen25vl} & 65.41 & 0.1295 & 0.1085 & 0.1155 \\
Qwen2.5-VL-72B~\citep{bai2025qwen25vl} & 77.60 & 0.2626 & 0.2337 & 0.2453 \\
\midrule
\multicolumn{5}{c}{\textit{Closed}} \\
\midrule
Gemini-2.5-Pro~\citep{comanici2025gemini25} & 85.65 & 0.2631 & 0.2192 & 0.2382 \\
GPT-o3~\citep{openai2025o3o4mini} & 85.60 & 0.3189 & 0.2690 & 0.2898 \\
GPT-5~\citep{openai2025gpt5} & 75.22 & 0.1790 & 0.1535 & 0.1658 \\
GPT-4o~\citep{openai2024gpt4o} & 87.76 & 0.3750 & 0.3487 & 0.3612 \\
\midrule
\multicolumn{5}{c}{\textit{Deepfake-Specialized}} \\
\midrule
AnomReasonor-7B~\citep{Tan2025Semantic} & 82.61 & 0.3684 & 0.3574 & 0.3613 \\
UniGenDet~\citep{Zhang2026Unigendet} & \textcolor{tabsecond}{\underline{88.59}} & 0.4729 & 0.3965 & 0.4234 \\
\midrule
\multicolumn{5}{c}{\textit{Ours}} \\
\midrule
Qwen3.5-9B~\citep{qwen2026qwen35} & 84.05 & 0.3268 & 0.2896 & 0.3063 \\
Qwen3.5-9B-LoRA-SFT~\citep{qwen2026qwen35,hu2022lora} & 77.83 & 0.5029 & 0.4794 & 0.4792 \\
Qwen3.5-9B-Iter1~\citep{qwen2026qwen35} & 85.72 & 0.5164 & 0.4909 & 0.4927 \\
Qwen3.5-9B-Iter2~\citep{qwen2026qwen35} & 87.58 & \textcolor{tabsecond}{\underline{0.5349}} & \textcolor{tabsecond}{\underline{0.5065}} & \textcolor{tabsecond}{\underline{0.5103}} \\
\rowcolor{softblue}
Qwen3.5-9B-Iter3~\citep{qwen2026qwen35} & \textcolor{tabbest}{\textbf{92.18}} & \textcolor{tabbest}{\textbf{0.5468}} & \textcolor{tabbest}{\textbf{0.5167}} & \textcolor{tabbest}{\textbf{0.5207}} \\
\bottomrule
\end{tabular}
}
\caption{Deepfake detection with explanation scoring on AnomReason-Deepfake~\citep{Tan2025Semantic}. Acc measures binary deepfake classification, while CSemAP reports classification-aware semantic quality for phenomenon (Phe), reasoning (Rea), and full matching. Red bold and blue underline mark the best and second-best values per column; the shaded row is our final model.}
\label{tab:anomreason_deepfake_results}
\end{table}

\subsection{Experimental Setup}

We evaluate \methodname{} on three externally curated benchmarks: DeepfakeJudge-Detect~\citep{Kuckreja2026Pixels}, a judge-style benchmark whose fake pool combines text-to-image generations with locally edited real photographs; AnomReason-Deepfake~\citep{Tan2025Semantic}, which scores both the verdict and the semantic quality of the accompanying explanation through classification-aware semantic AP (CSemAP); and Holmes-Set~\citep{Zhou2025Aigiholmes}, fully synthetic images from ten generator families that never appear in our training data. We anchor the evaluation on the first two, which are recent, far from saturated, and hard for different reasons: DeepfakeJudge-Detect draws its fakes from leaderboard-frontier commercial editors (Nano-Banana, SeedDream, Flux-Kontext, Qwen-Edit), while AnomReason-Deepfake screens its pool (Midjourney, SD3.5, FLUX) for photorealism and scores the explanation alongside the verdict. Holmes-Set is the mature fully-synthetic task and serves as an out-of-genre transfer probe. Baseline pools are benchmark-specific: each table compares against the strongest available systems of each kind, while UniGenDet and every \methodname{} stage appear in all three as a common reference. AnomReason-Deepfake and Holmes-Set are fully zero-shot; for DeepfakeJudge-Detect, only the SFT initialization uses the suite's designated training corpus. GenShield, which has no public release, appears only in the Holmes-Set comparison. Training configurations are listed in the technical appendix.

\subsection{Main Results}

\begin{figure*}[t!]
\centering
\includegraphics[width=0.92\textwidth]{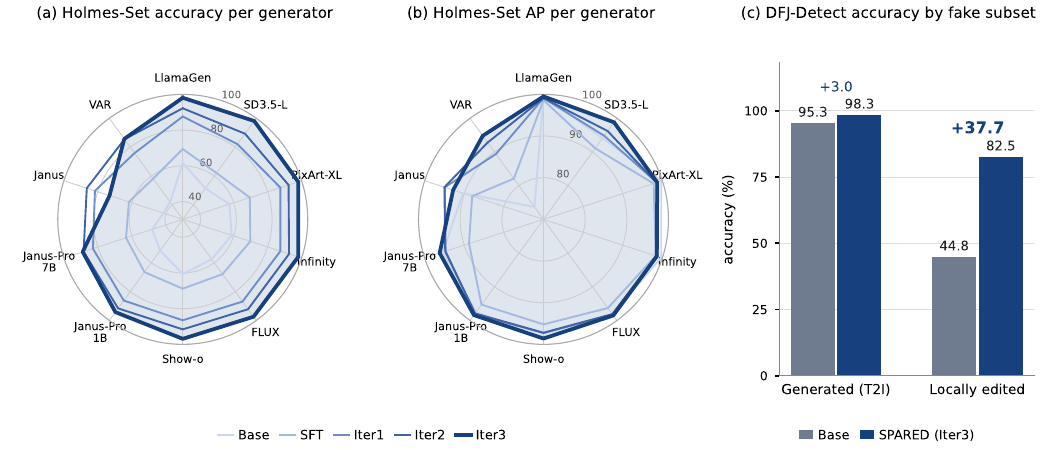}
\caption{(a, b) Per-generator accuracy and AP on Holmes-Set across training stages, with axes ordered by final-round accuracy: each adversarial round expands the accuracy polygon, and the final round's AP stays above 92 on every family. (c) DeepfakeJudge-Detect accuracy on the two fake subsets: the gain concentrates in the locally edited subset.}
\label{fig:analysis}
\end{figure*}

\paragraph{Judge-style detection (Table~\ref{tab:detect_revised}).}
Every adversarial round improves every aggregate metric on the primary benchmark: overall accuracy rises 69.6 $\rightarrow$ 72.1 $\rightarrow$ 76.0 $\rightarrow$ 79.5 from SFT through Iter3, with real recall (57.1 $\rightarrow$ 70.2) and fake recall (82.0 $\rightarrow$ 88.8) climbing together rather than trading off. At 9B parameters, the final model surpasses every non-reasoning MLLM we evaluate, including Qwen3-VL-235B (74.5), and every reasoning model up to 30B (77.5), trailing only Qwen3-VL-235B-Thinking (82.7), a model 26$\times$ its size. The per-class columns show where this margin comes from: strong closed models answer ``real'' almost unconditionally (96.6/95.8 real recall against 34.5/22.7 fake recall), while the synthetic-image specialist UniGenDet defaults the other way and answers ``fake'' on essentially every image (0.0/99.9); the same classification head reaches 99.2 on Holmes-Set (supplementary material), so its collapse here measures distribution shift, not a weak detector. Most open baselines sit near chance on one or both classes. \methodname{} and the 26$\times$-larger 235B-Thinking are the only two models holding both recalls above 70. The fake pool also splits by construction into a text-to-image and a locally edited subset. The generated subset is near saturation even for the base model (95.3, and 98.3 after training). The gain concentrates in the edited subset, where a small edit leaves most of the image real: accuracy rises from 44.8 to 82.5 (Figure~\ref{fig:analysis}c), matching a training pool built from such edits.

\paragraph{Verdict with reasoning (Table~\ref{tab:anomreason_deepfake_results}).}
On AnomReason-Deepfake, \methodname{} reaches 92.18 accuracy zero-shot, above every closed model we compare against (GPT-4o: 87.76), 3.6 points above the strongest baseline, UniGenDet (88.59), and 9.6 points above AnomReasonor (82.61), although the latter is fine-tuned on this benchmark's training data. Explanation quality moves the same way: CSemAP-Full reaches 0.5207, a 23\% relative improvement over the strongest baseline score (0.4234). SFT trades accuracy for the reasoning format (84.05 $\rightarrow$ 77.83 Acc, but 0.3063 $\rightarrow$ 0.4792 CSemAP); the subsequent GRPO rounds~\citep{shao2024deepseekmath}, whose reward never scores the explanation, then lift both metrics simultaneously and monotonically (85.72/0.4927 $\rightarrow$ 87.58/0.5103 $\rightarrow$ 92.18/0.5207). This is the behavior the Method predicts: once explanations can be reinforced only through the verdicts they lead to, explanation quality rises as a side effect of detection accuracy.

\paragraph{Generalization to unseen generators (Figure~\ref{fig:analysis}a; per-generator table in the supplementary material).}
Holmes-Set is the hardest transfer: the adversarial rounds train exclusively on locally edited photographs, yet must classify fully synthetic images from ten unseen generator families. Mean accuracy rises monotonically from 53.0 (base) through 65.7 (SFT) and 84.5 (Iter1) to 92.8 (Iter3) with zero in-domain data, within 2.8 points of AIGI-Holmes (95.6); Figure~\ref{fig:analysis}a shows the expansion per generator. Adding in-domain detection data to Iter3, a mix of the FakeClue corpus~\citep{Wen2025Spot} and the benchmark's own training set (Iter3+Holmes), reaches 97.2 mean accuracy and 99.9 mean AP, the best AP in the table, closing most of the distance to the strongest specialists, GenShield (98.8) and UniGenDet (99.2). The data alone does not explain this: the same mix given to the SFT model without adversarial rounds (SFT+Holmes) reaches 89.5, behind even the zero-shot Iter3. In-domain data compounds the adversarial rounds rather than substituting for them. This task is also close to saturation, with five detectors above 95, while the two anchored benchmarks still separate models sharply; UniGenDet reaches 99.2 here yet collapses on DeepfakeJudge-Detect, and \methodname{} holds both regimes (79.5 there, 92.8 zero-shot here).

\paragraph{Where the signal regresses.}
The ten-generator breakdown also locates the sharpest regression in the adversarial training signal: Janus accuracy drops from 86.5 (Iter2) to 73.1 (Iter3), even as every other family's accuracy improves and the mean rises from 90.2 to 92.8. This is not a loss of capacity: with in-domain data added, Janus recovers to 84.3 (Iter3+Holmes). Rather, the hard 0/1 fooling reward has no per-family difficulty control, so the family closest to the region the harder pool no longer exercises can regress even as the mean improves.
\begin{figure*}[t!]
\centering
\includegraphics[width=\textwidth]{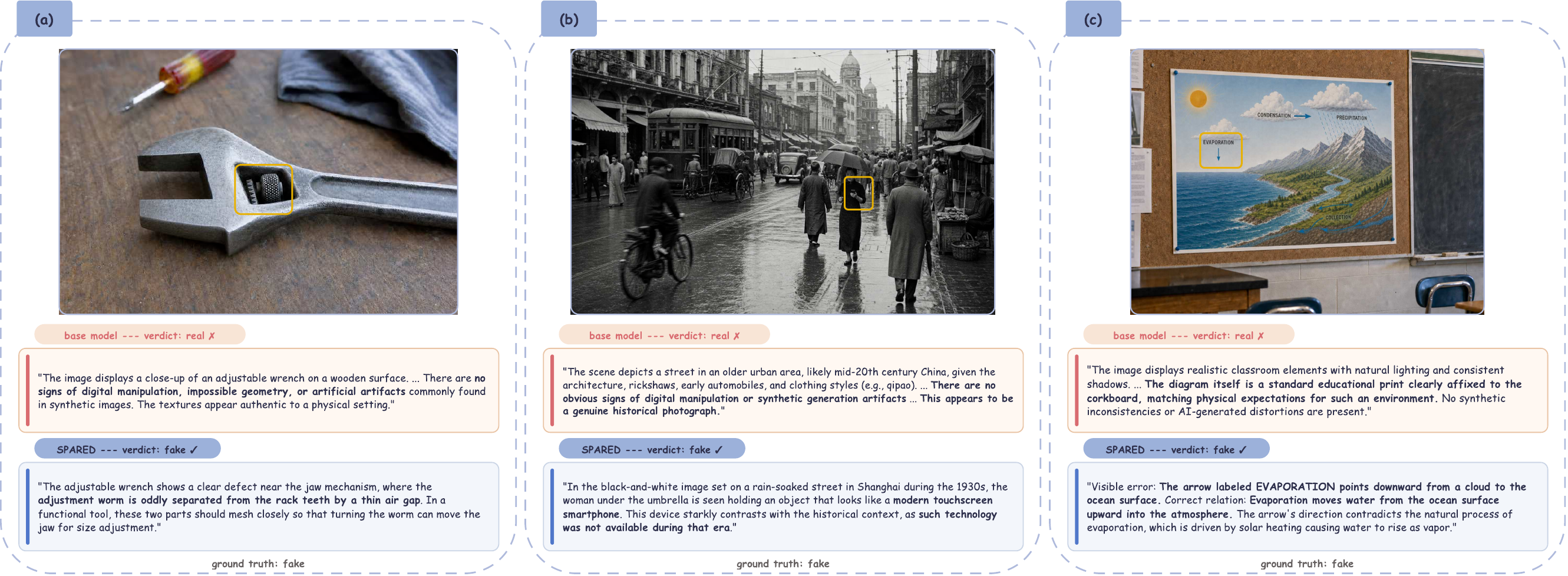}
\caption{Qualitative cases. (a) A mechanical violation: the base model judges the image real and explicitly reports no impossible geometry, while \methodname{} identifies the air gap separating the adjustment worm from the rack teeth and explains why the mechanism could not move the jaw. (b) A temporal violation: the base model accepts a 1930s Shanghai street scene as a genuine historical photograph, while \methodname{} finds the modern touchscreen smartphone in the woman's hand and rejects the image as anachronistic. (c) An information-visualization violation: the base model reads the classroom's water-cycle poster as a standard educational print, while \methodname{} locates the ``evaporation'' arrow pointing downward from cloud to ocean and states that the real process runs the other way.}
\label{fig:qualitative}
\end{figure*}

\subsection{Qualitative Analysis}

Figure~\ref{fig:qualitative} shows the trained detector on three generated images the base model accepts as real: a wrench whose adjustment worm cannot reach the rack teeth it is supposed to drive, a 1930s street scene holding a modern smartphone, and a classroom poster whose ``evaporation'' arrow points down into the ocean. Between them they instantiate the semantic-anomaly classes the Introduction argued remain detectable once low-level artifacts disappear: implausible geometry, and violations that take historical or scientific world knowledge to notice. In each case the verdict comes with evidence the reader can check against the image, the behavior the verdict-only reward was designed to elicit.

\subsection{Ablation Study}

\begin{table}[t!]
\centering
\renewcommand{\arraystretch}{1.0}
\resizebox{\columnwidth}{!}{
\begin{tabular}{lcccc}
\toprule
 & \multicolumn{2}{c}{\textbf{DFJ-Detect}} & \multicolumn{2}{c}{\textbf{AnomReason-DF}} \\
\cmidrule(lr){2-3}\cmidrule(lr){4-5}
\textbf{Variant} & Acc & F1 & Acc & CSemAP \\
\midrule
Iter1 (shared starting point) & 72.1 & 71.8 & 85.72 & 0.4927 \\
\ \ + static-pool GRPO ($4.7\times$ budget) & 74.3 & 74.3 & 88.00 & 0.4998 \\
\ \ + fresh-source GRPO, frozen attacker ($7.1\times$ budget) & 73.7 & 73.6 & 86.84 & 0.4990 \\
\ \ + one adversarial round, no PaCo gate & 66.7 & 63.8 & 81.12 & 0.4991 \\
\ \ + one adversarial round, unpaired reals & 73.2 & 73.0 & 85.18 & 0.5005 \\
\ \ + one adversarial round (Iter2) & 76.0 & 75.8 & 87.58 & 0.5103 \\
\rowcolor{softblue}
\ \ + adversarial rounds (Iter2 $\rightarrow$ Iter3) & \textcolor{tabbest}{\textbf{79.5}} & \textcolor{tabbest}{\textbf{79.3}} & \textcolor{tabbest}{\textbf{92.18}} & \textcolor{tabbest}{\textbf{0.5207}} \\
\bottomrule
\end{tabular}
}
\caption{Ablation on DeepfakeJudge-Detect (DFJ-Detect) and AnomReason-Deepfake (AnomReason-DF), all variants starting from the same Iter1 model: continued GRPO on the frozen round-one pool (static), on fresh-source pools regenerated by the frozen base editor (no attacker training), one adversarial round with the attacker's instruction-fidelity gate removed, one round on an unpaired pool (each fake's paired real replaced by a real from disjoint sources), one gated round (Iter2), and the full schedule (Iter2 $\rightarrow$ Iter3). Red bold marks the best value per column.}
\label{tab:ablation_static}
\end{table}

\paragraph{Adversarial rounds vs.\ continued static training (Table~\ref{tab:ablation_static}).}
If the adversarial rounds merely re-label the benefit of training longer, then freezing the loop after the first round and simply continuing GRPO on the same base-edited pool should match the full schedule. Table~\ref{tab:ablation_static} shows it does not: the static branch captures only 2.2 of the 7.4 points that the adversarial branch adds on DeepfakeJudge-Detect, and trails on every metric. The trajectory is stronger evidence than the endpoint: the static run peaks early, then declines; real recall collapses from the 60s into the 40s, consistent with a boundary drifting toward ``fake'', and explanation quality plateaus (CSemAP-Full never exceeds 0.5120). No static checkpoint approaches Iter3 on detection and explanation jointly, whereas the adversarial rounds lift both monotonically. A second control rules out fresh data alone: regenerating the pool each round from freshly sampled sources with the \emph{frozen} base editor, from the same Iter1 model, peaks at 73.7 on the primary benchmark and oscillates below that thereafter, despite consuming more total budget than the entire schedule. More optimization does not close the gap, and neither does fresher data; the gains require an attacker that keeps learning to fool the current defender.

\paragraph{Gating the attacker's reward (Table~\ref{tab:ablation_static}).}
We repeat the first attacker round with the gate removed, rewarding fooling alone. At matched budget its behavior drifts as the Method predicts: it passes the instruction-fidelity check 9.6 points less often, its median edit magnitude shrinks by 27\%, and it fools the frozen defender slightly more, winning more while editing less. Training a defender round on this pool turns drift into damage: instead of reproducing Iter2's gain, the round falls below its starting point (66.7 vs.\ Iter1's 72.1 on DeepfakeJudge-Detect; 81.12 vs.\ 85.72 on AnomReason-Deepfake). The failure is diagnostic: real recall collapses to 38.2 while fake recall inflates to 95.1, consistent with a pool whose barely-edited ``fakes'' teach the defender that realistic-looking images are fake, which is exactly what the gate exists to prevent.

\paragraph{Unpairing the training pool (Table~\ref{tab:ablation_static}).}
The last lock is the pairing itself. If the pairing were cosmetic, replacing the reals with photographs from disjoint sources (same fakes, same budget) should not matter. Instead, most of the round's benefit disappears: the unpaired round gains 1.1 points on DeepfakeJudge-Detect where the paired round (Iter2) gains 3.9, and on AnomReason-Deepfake it lands below even its starting point (85.18 vs.\ Iter1's 85.72, against Iter2's 87.58). Part of the training signal is spent on provenance cues that do not transfer, the dataset-bias shortcut the pairing exists to close.
\section{Conclusion}

We presented \methodname{}, a decoupled attacker--defender reinforcement-learning loop for reasoning-based AI-generated image detection: a diffusion image editor forges paired fakes under an instruction-fidelity gate, and a reasoning MLLM learns to expose them under a verdict-only reward. Once every channel connecting the two models is shortcut-proof, adversarial pressure converts into capability: the detector improves monotonically across rounds and transfers to three external benchmarks, two of them zero-shot, and neither longer training, fresher data, nor in-domain data substitutes for an evolving attacker. Its main limitation is the hard 0/1 fooling reward: without per-family difficulty control, an isolated per-generator regression persists even as the means rise. A graded difficulty signal is the natural next step.

\bibliography{aaai2027}

\end{document}